\documentclass[letterpaper, 10 pt, conference]{ieeeconf}  % Comment this line out if you need a4paper

\IEEEoverridecommandlockouts                              % This command is only needed if 
\usepackage{amsmath}
\usepackage{amssymb}
\usepackage{filecontents}
\usepackage{float}
\usepackage{pgfplots}
\usepackage{subcaption}
\usepackage{tabularx}
\usepackage{tikz}
\usetikzlibrary{arrows}

\title{\LARGE \bf Beelines: Motion Prediction Metrics for Self-Driving Safety and Comfort}
\author{Skanda Shridhar, Yuhang Ma, Tara Stentz, Zhengdi Shen, Galen Clark Haynes, Neil Traft\thanks{The authors are with Uber Advanced Technologies Group, PA, USA.}}
\date{October 2020}

\begin{document}

\maketitle

%%%%%%%%%%%%%%%%%%%%%%%%%%%%%%%%%%%%%%%%%%%%%%%%%%%%%%%%%%%%%%%%%%%%%%%%%%%%%%%%
\begin{abstract}
The commonly used metrics for motion prediction do not correlate well with a self-driving vehicle's system-level performance. The most common metrics are \textit{average displacement error (ADE)} and \textit{final displacement error (FDE)}, which omit many features, making them poor self-driving performance indicators.  Since high-fidelity simulations and track testing can be resource-intensive, the use of prediction metrics better correlated with full-system behavior allows for swifter iteration cycles. In this paper, we offer a conceptual framework for prediction evaluation highly specific to self-driving. We propose two complementary metrics that quantify the effects of motion prediction on safety (related to recall) and comfort (related to precision). Using a simulator, we demonstrate that our safety metric has a significantly better signal-to-noise ratio than displacement error in identifying unsafe events.
\end{abstract}

%%%%%%%%%%%%%%%%%%%%%%%%%%%%%%%%%%%%%%%%%%%%%%%%%%%%%%%%%%%%%%%%%%%%%%%%%%%%%%%%
\section{Introduction}

\begin{figure}[h]
\centering
\begin{subfigure}{.4\textwidth}
  \centering
  \captionsetup{justification=centering}
  \includegraphics[width=.8\textwidth]{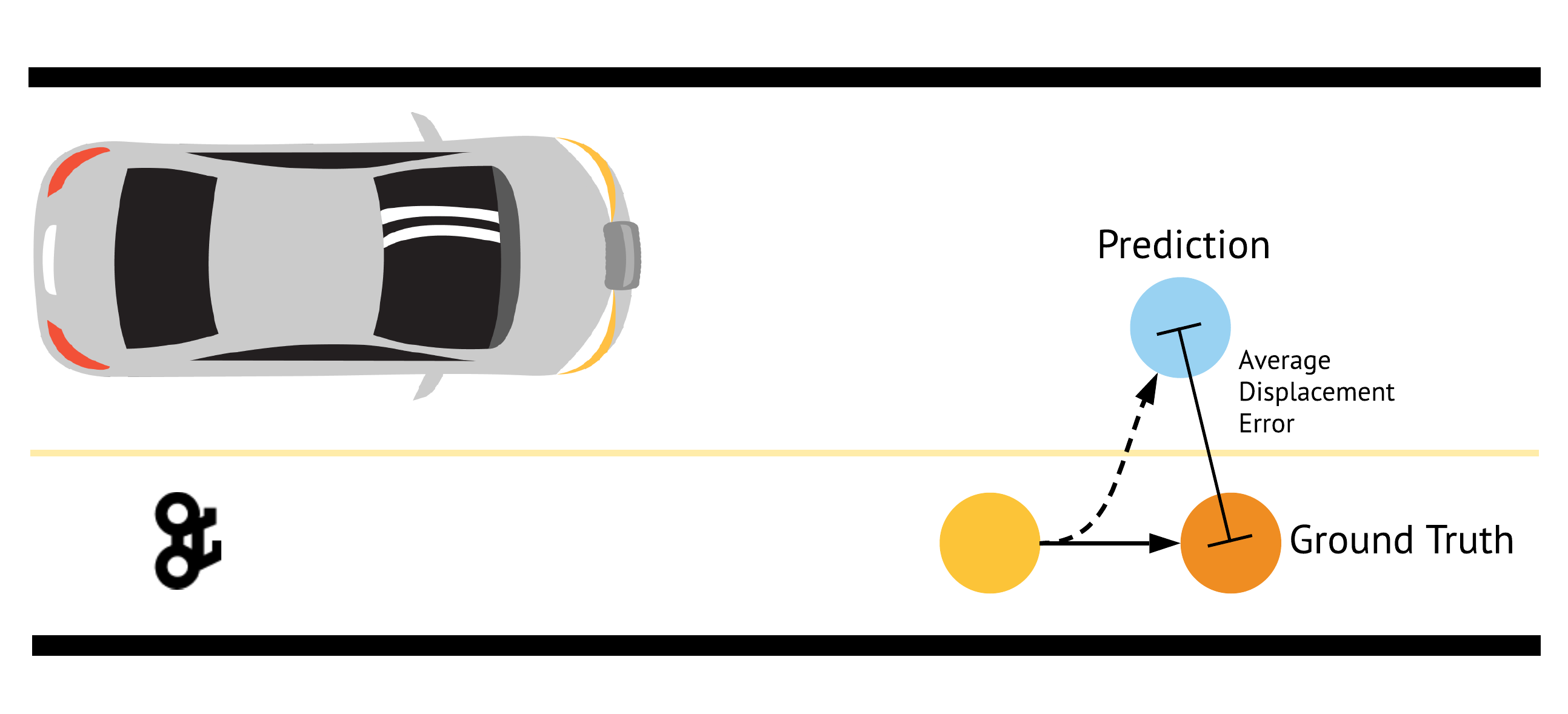}
  \caption{The future prediction (blue) is in the vehicle's path while the future ground-truth (orange) is not.}
  \label{fig:fp}
\end{subfigure}
\begin{subfigure}{.4\textwidth}
  \centering
  \captionsetup{justification=centering}
  \includegraphics[width=.8\textwidth]{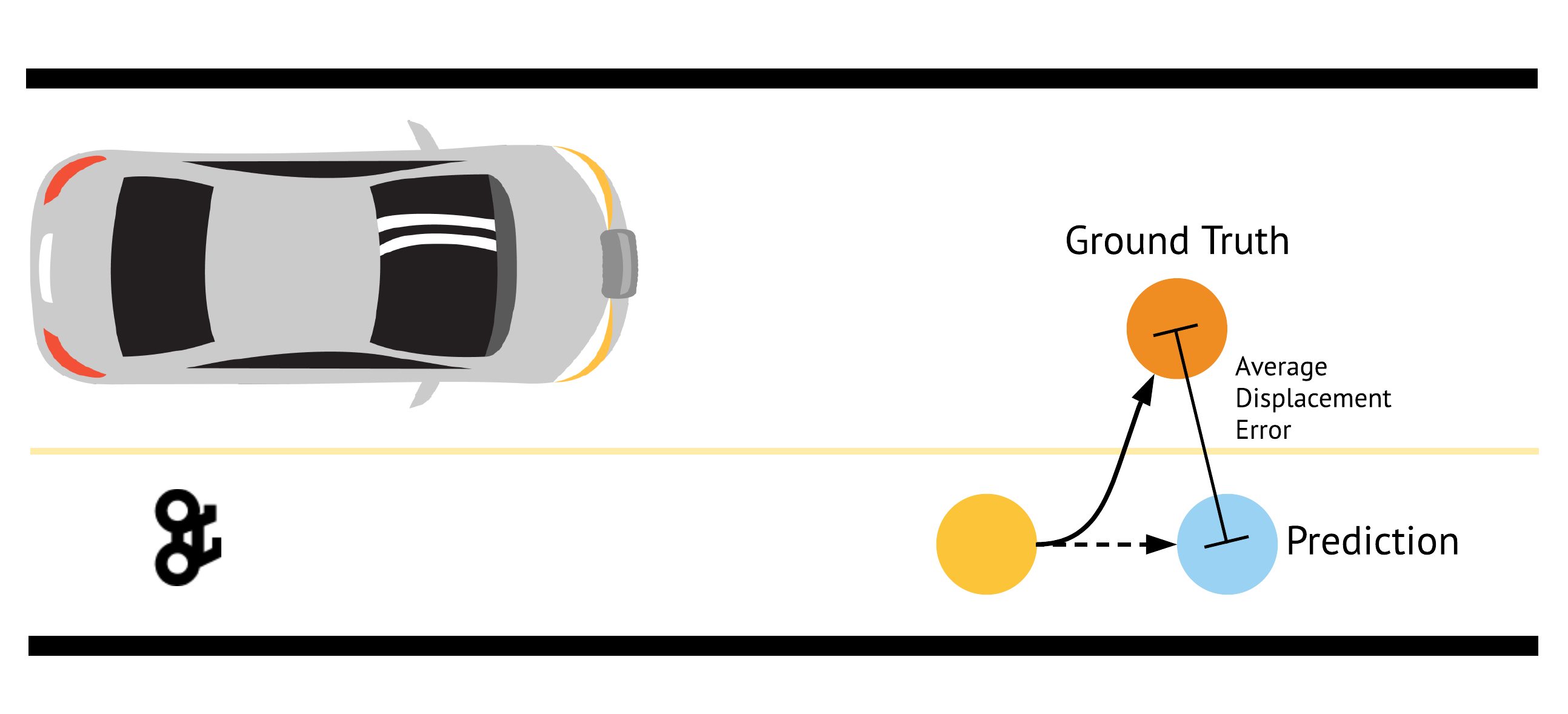}
  \caption{The future ground-truth (orange) is in the vehicle's path while the future prediction (blue) is not.}
  \label{fig:fn}
\end{subfigure}
\caption{Two examples of prediction errors, with identical displacement error. The bottom is more of a safety concern.}
\label{fig:fp_fn}
\end{figure}

In self-driving systems, actor detection and motion prediction are essential. Most methods train machine-learned models to predict trajectories or occupancy maps. These models are usually trained on variants of the L2 or cross-entropy loss and evaluated with metrics such as \textit{average displacement error (ADE)} and \textit{final displacement error (FDE)}. However, consider the shortcoming illustrated by Fig.~\ref{fig:fp_fn}, which shows two simple scenarios with identical displacement error, but in which the error in Fig.~\ref{fig:fn} is more severe because the vehicle has not predicted the object's future entry into its path. 

As a result, evaluating predictions' impact on on-road behavior usually entails full-system simulation and track testing. Both are expensive and require a fully-capable self-driving platform. Moreover, system factors beyond the actor motion prediction method (for instance, the particular planning algorithm) can influence results, making it difficult to isolate the predictions' impact. A desirable motion prediction metric would anticipate its system-level impact while remaining agnostic of the broader system.

The \textit{safety} impact of predictions is related to recall. When we forecast an actor's future motion, we predict what parts of the world they will occupy. It follows then that \emph{recall} is the percentage of the actor's future occupied space that has been covered by the predictions made. Low recall suggests that a self-driving vehicle does not correctly anticipate the actor's future occupancy regions, a safety concern.

Conversely, the \textit{comfort} impact of predictions is related to precision. Precision is the fraction of the space predicted occupied that does end up as occupied. Low precision suggests that the system is over-predicting, and this will make the ego-vehicle brake unnecessarily. In extreme cases, it may be unable to move.

This work proposes a metrics framework that directly connects precision and recall to the self-driving specific concerns of ride comfort and safety. It addresses many flawed properties of metrics that evaluate predictions in isolation (such as ADE), including:

\begin{itemize}
  \item Insensitivity to Object Shape and Orientation Error.
  
  \textit{Solution:} Since we deal with occupied space, we account for the position, orientation, shape, and uncertainty error.
  \item Reporting a Single Horizon or Aggregating Over All. 
  
  \textit{Solution:} Since we evaluate predictions in relation to ego-vehicle spatiotemporal maneuvers, we consider future horizons holistically.
  \item Insensitivity to Ego-Vehicle (as in Fig.~\ref{fig:fp_fn}).
  
  \textit{Solution:} We grade safety based on the predictions' ability to ``protect'' ground-truth ``exposed'' to the ego-vehicle, and comfort based on the extent to which predictions deprive the ego-vehicle of maneuvers.
\end{itemize}

While system-level evaluation (e.g., simulation) addresses the same concerns, it has shortcomings of its own, which we seek to overcome:
\begin{itemize}
  \item Inability to Isolate Detection and Prediction Impact from Broader System.
  
  \textit{Solution:} Since we use fixed ego-vehicle maneuvers, we anticipate the system-level impact. Since these are system-agnostic, we isolate detection and prediction impact from the broader system.
  \item One Scenario Per Simulation.
  
  \textit{Solution:} We evaluate 1000s of counterfactual ego-vehicle behaviors and marginalize over these in a single pass.
    \item Difficulty Backpropagating Error Gradients.
    
    \textit{Solution:} Our approach lends itself to a GPU vectorized implementation through which we can backpropagate error gradients.
\end{itemize}

We show that our method is better than displacement error at marking prediction failures that lead to unsafe events in a simulator. 

%%%%%%%%%%%%%%%%%%%%%%%%%%%%%%%%%%%%%%%%%%%%%%%%%%%%%%%%%%%%%%%%%%%%%%%%%%%%%%%%
\section{Related Work}

The most widely used metrics, ADE and FDE, were proposed in \cite{pellegrini2009} and popularized by the TrajNet benchmark \cite{trajnet, red-predictor}. Since it is unclear what the most important prediction horizon is,  works such as  \cite{djuric2018, becker2019rnn} report these metrics at multiple horizons. The nuScenes Prediction Challenge \cite{nuscenes-challenge} and Argoverse Motion Forecasting Challenge \cite{argoverse-challenge} use the \emph{minADE} variant, also referred to as ``oracle error'' \cite{desire}, which takes the minimum error over the top $k$ trajectories. Many multimodal prediction models \cite{cui2019, desire, rupprecht2017, safecritic} use this, but it does not penalize false-positive modes, and we must consider it alongside ADE for a complete picture.

The Lyft Motion Prediction Competition \cite{lyft-challenge} and other works \cite{precog, schulz2018interaction, multipath} output a 2D probability distribution in the state space, and so can replace ADE with negative log-likelihood (NLL).

All these metrics have two drawbacks: 1) They only consider a single point on an object and do not account for orientation, shape, or relevance to the ego-vehicle, and 2) They require \emph{ground-truth matching} to use in actual practice. Since we define them on a pair of points (prediction and the ground-truth), we must first match each predicted actor with a ground-truth label. If we cannot associate a prediction (false-positive), we must ignore it and likewise for ground-truth. To capture these omissions, we must rely on other metrics, and this lack of comprehensiveness is a drawback.

Methods like ours, which evaluate an occupancy representation instead of trajectories, can transcend this limitation \cite{jain2019, chauffeurnet, faf}. Their evaluation is usually done with NLL or cross-entropy loss. Lacking awareness of the ego-vehicle, this suffers the same fundamental drawback highlighted in Fig.~\ref{fig:fp_fn}.

Jain et al. (2019) \cite{jain2019} also report \emph{safety-sensitive} recall, which takes the drivable region into account and is therefore closely related to our work. While this gives some safety-related feedback, it does not incorporate comfort nor relevance to the ego-vehicle. Our framework addresses these factors in full.

%%%%%%%%%%%%%%%%%%%%%%%%%%%%%%%%%%%%%%%%%%%%%%%%%%%%%%%%%%%%%%%%%%%%%%%%%%%%%%%%
\section{Method Outline}

We evaluate actor predictions in relation to a set of fixed maneuvers the ego-vehicle may perform, the \emph{ego-vehicle trajectories}. We frame the safety problem by asking: Given a distribution of ego-vehicle trajectories, how well do the predictions protect real-world objects?  The comfort problem is similarly posed: Given the same distribution of maneuvers, what fraction are needlessly blocked? To quantify the answers, we:
\begin{itemize}
  \item Generate a set of fixed ego-vehicle trajectories to approximate the full set of dynamically feasible maneuvers. (Section \ref{s:path-relative})
  \item Given the prediction output, for each ego-vehicle trajectory,
    \begin{itemize}
      \item Compute the likelihood that any exposed objects along that trajectory are not ``blocked'' by predictions. (Section \ref{s:defns})
      \item Compute the likelihood that any reachable free space along that trajectory is ``blocked'' by predictions. (Section \ref{s:defns})
    \end{itemize}
    \item Marginalize across all trajectories. (Section \ref{s:beeline-agg})
\end{itemize}

%%%%%%%%%%%%%%%%%%%%%%%%%%%%%%%%%%%%%%%%%%%%%%%%%%%%%%%%%%%%%%%%%%%%%%%%%%%%%%%%
\section{Protection and Exposure} \label{s:defns}
\begin{figure*}[h]
\centering
\captionsetup{justification=centering}
\includegraphics[width=.9\linewidth]{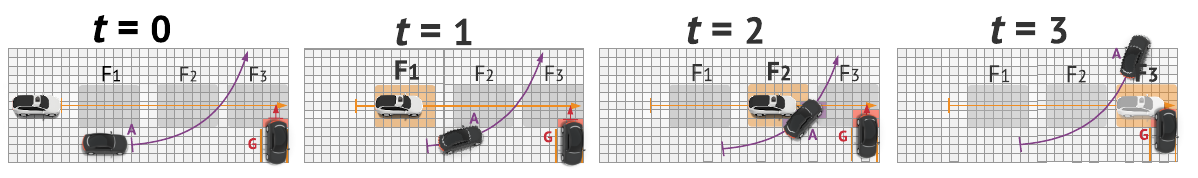}
\caption{At $t=0$, a fixed ego-vehicle trajectory is shown as a sequence of its future footprints, $F_1, F_2,$ and $F_3$. At $t=1$, the ego-vehicle occupies $F_1$. At $t=2$, it occupies $F_2$ and overlaps $A$'s space. It cannot reach $F_3$ and encounter $G$ at $t=3$ (hence shown faded) since it must re-plan to stop for $A$, so $A$ blocks $G$ along this specific ego-vehicle trajectory.} 
\label{fig:blockage}
\end{figure*}
Let us first assume that we have a set of ego-vehicle trajectories available (we discuss generating this in more detail in Section VI). We can think of an ego-vehicle trajectory as a region of fixed area (the ego-vehicle's body) moving through time. At a given time, we refer to the region as the ego-vehicle's \textit{footprint}. Figure \ref{fig:blockage} represents the ego-vehicle's trajectory as a sequence of its footprints. It also represents ground-truth actors and predictions with their footprints in the same spatiotemporal volume.

When executing a fixed trajectory, if one region prevents the ego-vehicle from accessing another by requiring it to stop, the former \textit{blocks} the latter. For instance, if queued behind a line of cars, the car immediately in front blocks all subsequent cars. In Fig. \ref{fig:blockage}, actor $G$ is blocked by actor $A$ at the earlier time. Predictions have the effect of blocking real-world actors since they induce the ego-vehicle to stop.

%%%%%%%%%%%%%%%%%%%%%%%%%%%%%%%%%%%%%%%%%%%%%%%%%%%%%%%%%%%%%%%%%%%%%%%%%%%%%%%%
\subsection{Protection of Space}
The \emph{protection} of a space is the probability that a prediction blocks that space.

In the spatiotemporal grid setting of Fig. \ref{fig:blockage}, let $P_{pred}(x)$ be the predicted probability that a cell $x$ is occupied. The system either produces this directly or produces an output from which this can be derived. Given the ego-vehicle trajectory's footprint at time $t$, denoted $F_t$, the predicted probability that at least one cell in the footprint is occupied is given by,
\begin{equation}
P_{pred}(F_t \textrm{ is Occupied}) = 1 - \prod_{x \in F_{t}} [1 - P_{pred}(x)]
\end{equation}
At time $H$, the probability of $F_H$ being unprotected is the joint probability of no predicted occupancy in any footprint up to $H$: 
\begin{equation} \label{eq:protection}
	P(F_{H}\textrm{ is Unprotected}) = \prod_{t=t_1}^{H} [1 - P_{pred}(F_t\textrm{ is Occupied})] 
\end{equation} 
where $t_1$ is the time of the first footprint of interest. Note that we have assumed independence across spatiotemporal cells.

If $A$ in Fig. \ref{fig:blockage} is a prediction, and $G$ is a ground-truth object. Then,
\begin{multline*}
 P(F_{3}\textrm{ is Unprotected}) = [1 - P_{pred}(F_1\textrm{ is Occupied})]  \\
 \cdot [1 - P_{pred}(F_2\textrm{ is Occupied})]  \cdot [1 - P_{pred}(F_3\textrm{ is Occupied})]\\
 = [1 - (1 - \prod_{x \in F_{1}} [1 - P_{pred}(x)])] \cdot [1-(1 - \prod_{x \in F_{2}} [1 - P_{pred}(x)])] \\
 \cdot [1-(1 - \prod_{x \in F_{3}} [1 - P_{pred}(x)])] \\
 = 1 \cdot 0 \cdot 1 = 0 \\
\end{multline*}
which quantifies our intuition that $A$ blocks $G$, and therefore $G$ cannot be unprotected.

%%%%%%%%%%%%%%%%%%%%%%%%%%%%%%%%%%%%%%%%%%%%%%%%%%%%%%%%%%%%%%%%%%%%%%%%%%%%%%%%
\subsection{Exposure of Space}
The \emph{exposure} of a space (such as an ego-vehicle footprint) is the probability that no ground-truth blocks that space. Let $P_{gt}(x)$ be the ground-truth probability that a cell $x$ is occupied.  We obtain this, for instance, by converting available ground-truth object polygons into a spatiotemporal occupancy grid. Then, given a footprint $F_t$ at time $t$ of a specific ego-vehicle trajectory, 
\begin{equation}
P_{gt}(F_t \textrm{ is Occupied}) = 1 - \prod_{x \in F_{t}} [1 - P_{gt}(x)]
\end{equation}
At time $H$, 
\begin{equation}
P(F_{H} \textrm{ is Exposed}) = \prod_{t=t_1}^{H-1}[1 - P_{gt}(F_t \textrm{ is Occupied})] \label{eq:exposed}
\end{equation}
where $t_1$ is the first footprint's time. Note the use of $H-1$ rather than $H$ as in Eq. \ref{eq:protection}, which prevents us from crediting a ground-truth region for blocking itself.
Real-world objects overlapping high-exposure footprints are vulnerable, and predictions should prioritize blocking them.

%%%%%%%%%%%%%%%%%%%%%%%%%%%%%%%%%%%%%%%%%%%%%%%%%%%%%%%%%%%%%%%%%%%%%%%%%%%%%%%%
\section{Quantifying Safety and Comfort Impact}  \label{s:p_lambda} \label{s:beeline-agg}
\subsection{Quantifying Safety Across Ego-Vehicle Trajectories}
Assume $B$ is the set of fixed ego-vehicle trajectories. Let $T$ be the set of valid trajectory horizons $\{0, 1, ... T_{max}\}$. We define a risky event $\lambda$ as occurring when the system does not protect exposed real-world occupancy. We quantify this risk by finding the fraction of the exposed, reachable space that is occupied but unprotected:

\begin{equation}\label{eq:p_lambda}
P(\lambda) = \frac{\sum_{b,t \in B\times T} P(F_{t}^{b}\textrm{ is Reached}) d(F_{t}^{b})}{\sum_{b,t \in B\times T} P(F_{t}^{b}\textrm{ is Reached}) e(F_{t}^{b})} 
\end{equation}

where,
\begin{align*}
 d(F_{t}^{b}) &= P(F_{t}^{b} \textrm{ is Unprotected}) \cdot \\ 
 & P_{gt}(F_{t}^{b} \textrm{ is Occupied}) \cdot 
 	P(F_{t}^{b} \textrm{ is Exposed}) \\
e(F_{t}^{b}) &= P(F_{t}^{b} \textrm{ is Exposed} )
\end{align*}
$P(F_{t}^{b}\textrm{ is Reached})$ is the probability of the ego-vehicle occupying $F_{t}^{b}$, its footprint at time $t$ when executing trajectory $b$. 

Regions where $e(F_{t}^{b})=0$ represent footprints inaccessible to the ego-vehicle because they are blocked by ground-truth, and these are excluded from the risk calculation.

\begin{figure*}[h]
\centering
\begin{subfigure}{.15\textwidth}
  \centering
  \captionsetup{justification=centering}
  \includegraphics[width=\textwidth]{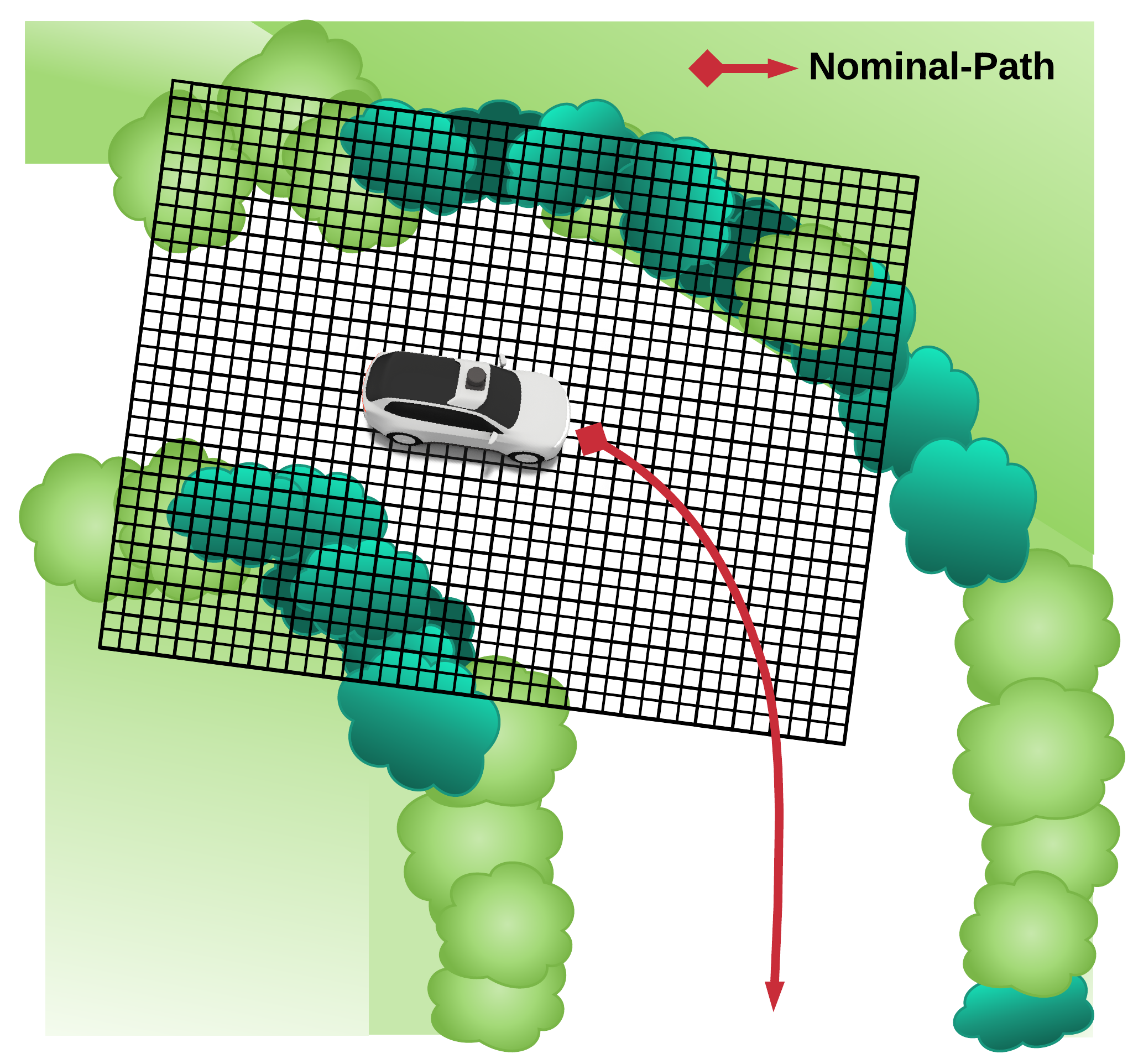}
  \caption{A naive grid. Much of it covers area that is irrelevant to the ego-vehicle's concerns.}
  \label{fig:naive_grid}
\end{subfigure}
\begin{subfigure}{.15\textwidth}
  \centering
  \captionsetup{justification=centering}
  \includegraphics[width=\textwidth]{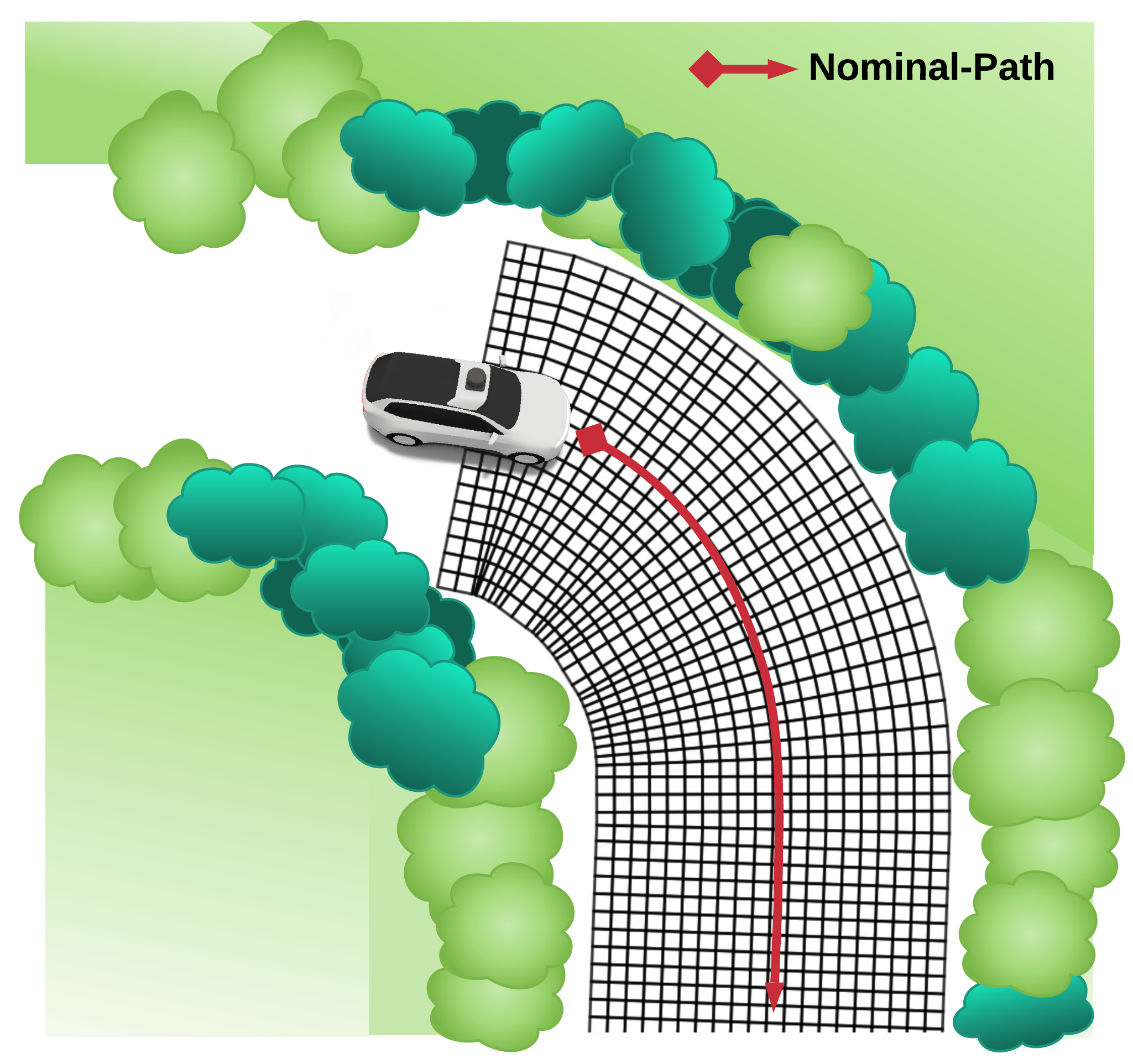}
  \caption{A curvilinear mesh that respects road-geometry. Produced by using the nominal-path's inverse path-relative transform.}
  \label{fig:sro_grid}
\end{subfigure}
\begin{subfigure}{.5\textwidth}
  \centering
  \captionsetup{justification=centering}
  \includegraphics[width=\textwidth]{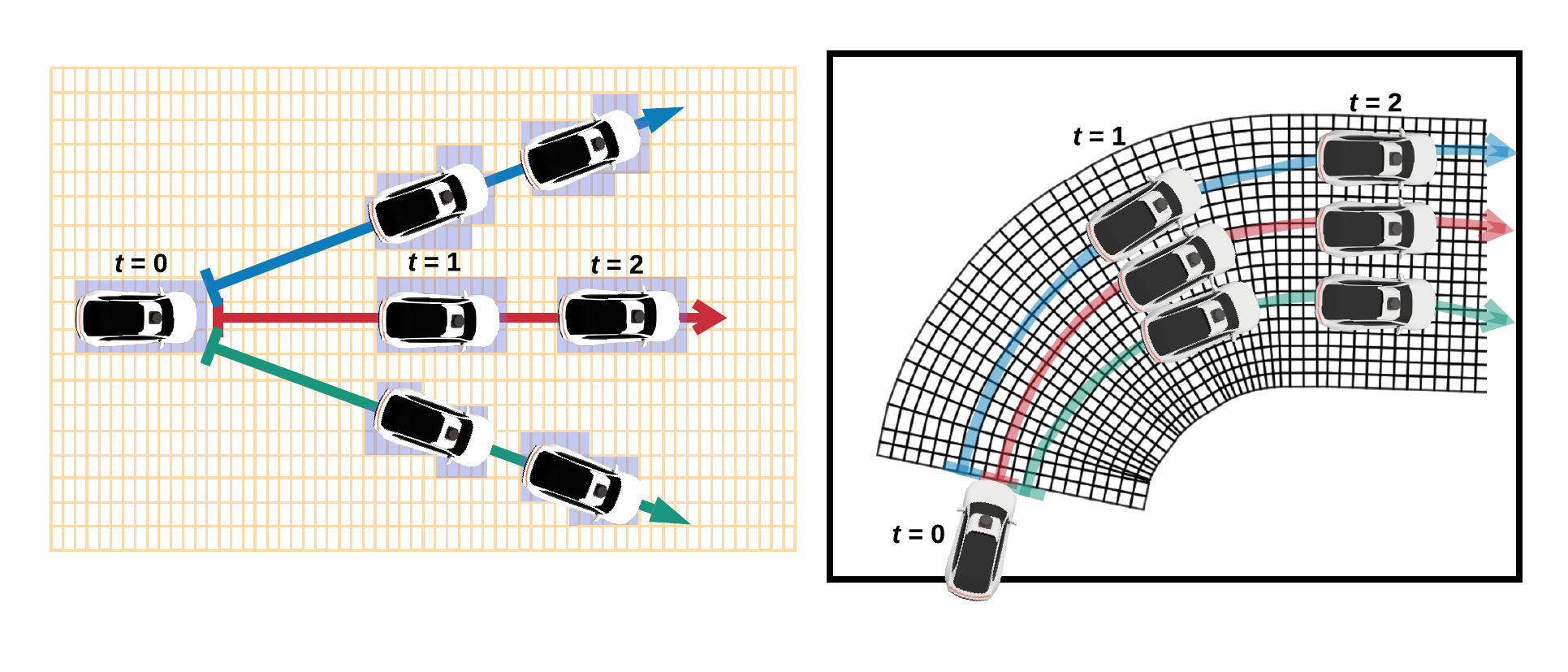}
  \caption{Beelines are produced in the path-relative frame (left). When transformed to real-world coordinates (right) using a path, they form spatially reasonable trajectories.}
  \label{fig:beelines_pr_to_abs}
\end{subfigure}
\caption{Given a path, we can use the path-relative transformation to apply the same set of beelines to any scene.}
\label{fig:sro_intro}
\end{figure*}

If $A$ in Fig. \ref{fig:blockage} is a prediction and $G$ is ground-truth, and there is only one ego-vehicle trajectory, and the footprints at all times are equally likely, i.e. $P(F_t\textrm{ is Reached}) = 1/3$,
\begin{multline*}
P(\lambda) = \frac{\frac{1}{3} \cdot 1 \cdot 0 \cdot 1   + \frac{1}{3} \cdot 0 \cdot 0 \cdot 1 + \frac{1}{3} \cdot 0 \cdot 1 \cdot 1}{\frac{1}{3} \cdot 1 + \frac{1}{3} \cdot 1 + \frac{1}{3} \cdot 1} 
 = 0
\end{multline*}
which captures the intuition of $A$ protecting $G$. If however, $A$ is ground-truth,
\begin{multline*}
P(\lambda) = \frac{\frac{1}{3} \cdot 1 \cdot 0 \cdot 1   + \frac{1}{3} \cdot 1 \cdot 1 \cdot 1 + \frac{1}{3} \cdot 1 \cdot 1 \cdot 0}{\frac{1}{3} \cdot 1 + \frac{1}{3} \cdot 1 + \frac{1}{3} \cdot 0} 
 = 0.5
\end{multline*}
This is because two equally likely footprints ($F_1$, $F_2$) are ``reachable" to the ego-vehicle and one of them $F_2$ involves an unsafe event, i.e. colliding with unprotected ground-truth $A$. Since $F_3$ is blocked by $A$ it is excluded.

If interested in a particular actor, we may marginalize the numerator of Eq. \ref{eq:p_lambda} over the subset of ego-vehicle trajectory footprints that intercept the actor. We term this actor-specific quantity $P(\lambda_{actor})$.

The expression $e(F_{t}^{b})$, when marginalized in Eq. \ref{eq:p_lambda}, represents the probability apportioned to ego-vehicle states not blocked by ground-truth. If we modify it as follows,
\begin{equation} \label{eq:modified_eFt}
e'(F_{t}^{b}) = P(F_{t}^{b} \textrm{ is Exposed} ) P(F_{t}^{b} \textrm{ is Unprotected})
\end{equation}
we also exclude states blocked by predictions. In addition to safety, this penalizes constricting the ego-vehicle's space. The next section discusses isolating comfort impact.

%%%%%%%%%%%%%%%%%%%%%%%%%%%%%%%%%%%%%%%%%%%%%%%%%%%%%%%%%%%%%%%%%%%%%%%%%%%%%%%%
\subsection{Quantifying Comfort Across Ego-Vehicle Trajectories}
A comfort violation $\zeta$ occurs when the predictions block free footprints. We capture this by finding the fraction of exposed, reachable space that is unoccupied but blocked by predictions:
\begin{equation}
P(\zeta) = \frac{\sum_{b,t \in B\times T} P(F_{t}^{b}\textrm{ is Reached}) h(F_{t}^{b})}{\sum_{b,t \in B\times T} P(F_{t}^{b}\textrm{ is Reached}) g(F_{t}^{b})} 
\label{eq:p_zeta}
\end{equation}
where,
\begin{align*}
 h(F_{t}^{b}) = (1 - P(F_{t}^{b} \textrm{ is Unprotected})) \cdot \\(1-P_{gt}(F_{t}^{b} \textrm{ is Occupied})) \cdot 
 	\\ P(F_{t}^{b} \textrm{ is Exposed}) \\
g(F_{t}^{b}) = (1-P_{gt}(F_{t}^{b} \textrm{ is Occupied})) \cdot \\P(F_{t}^{b} \textrm{ is Exposed} )
\end{align*}
Regions where $g(F_{t}^{b}) = 0$ represent footprints that are blocked by real-world obstacles and inaccessible to the ego-vehicle, and we skip the calculation for these regions. In Fig. \ref{fig:blockage} if $A$ is a prediction and $G$ is ground-truth,
\begin{multline*}
P(\zeta) = \frac{\frac{1}{3} \cdot 0 \cdot 1 \cdot 1  + \frac{1}{3} \cdot 1 \cdot  1 \cdot 1 + \frac{1}{3} \cdot 1 \cdot 0 \cdot 1}{\frac{1}{3} \cdot 1 \cdot 1  + \frac{1}{3} \cdot 1 \cdot 1  + \frac{1}{3} \cdot 0 \cdot 1 } 
 = 0.5
\end{multline*}
This score results from the prediction arriving a timestep too early. Of the two available footprints $F_1$, $F_2$ ($G$ occupies $F_3$), $A$ blocks one, and therefore deprives the ego-vehicle of half its maneuverable space. 

%%%%%%%%%%%%%%%%%%%%%%%%%%%%%%%%%%%%%%%%%%%%%%%%%%%%%%%%%%%%%%%%%%%%%%%%%%%%%%%%
\section{Implementation of Method} \label{s:path-relative}

If available, a motion planner may be used to produce the fixed ego-vehicle trajectories. However, this may be an excessive computational burden. Therefore we present a simple, performant approximation of a motion planner that uses the path-relative transformation.

\subsection{Path-Relative Space}
A Cartesian grid will frequently result in the situation of Fig. \ref{fig:naive_grid}, where much of the grid is irrelevant. A solution that adapts to the intended travel path, as shown in Fig. \ref{fig:sro_grid}, is preferable. The path-relative transformation enables this. 

To transform a point $p_{abs} = (x, y)$ from real-world coordinates into path-relative, $p_{pr} = (a, c)$, we first define the  origin as the closest point on the path to the ego-vehicle's position. Next, we find $p_{proj}$, the closest point on the path  to $p_{abs}$. The along-path distance from the origin to $p_{proj}$ gives the along-track coordinate, $a$. The distance between $p_{abs}$ and $p_{proj}$ along the normal to the path at $p_{proj}$ gives the cross-track coordinate, $c$.

We define the inverse transformation similarly; given $(a, c)$ in path-relative coordinates, we advance a distance $a$ along the nominal-path. From there, we shift $c$ in the normal direction to get $p_{abs}$.

\subsection{Beelines} \label{s:beelines}
In path-relative space, we produce trajectories spanning a range of initial heading angles with constant acceleration over their lifetime. We call them \textit{beelines}. When transformed into real-world coordinates, these beelines are spatially appropriate for the scene, as in  Fig. \ref{fig:beelines_pr_to_abs}. Once generated, we can apply the set of beelines to any scene, given a nominal-path.

\subsection{Probability of an Ego-Vehicle Footprint} \label{s:beeline-likelihood}
We define independent distributions $f_{\theta}(\theta)$ and $f_{a}(a)$ over the initial heading $\theta$ and acceleration $a$ of the ego-vehicle in path-relative space. Given a starting velocity $v_{i}$ and a footprint time $t$, since $r = v_{i}t + \frac{1}{2}at^{2}$, we can define a density of possible footprint centers in polar coordinates, using $f_{\theta}(\theta)$ and applying a change of variables to $f_a(a)$,
\begin{equation}
f(r, \theta \mid v_{i}, t) = \frac{2}{t^{2}} f_{a} \bigg(\frac{2(r - v_i t)}{t^2} \bigg) f_{\theta}(\theta)
\end{equation}
where we assume constant acceleration and heading. Transforming to Cartesian coordinates, and distributing the density uniformly over time up to a maximum horizon $T_{max}$, 
\begin{equation}
  f(x,y,t \mid v_{i}) = 
  \begin{cases}
    \frac{2}{T_{max}t^{2}} \frac{f_{a} \big( \frac{ 2 (\sqrt{x^{2} + y^{2}} - v_i t)}{t^2} \big) }{\sqrt{x^{2} + y^{2}}} f_{\theta}(tan^{-1}(y/x)), \\ t \in (0, T_{max}] \\
    0, \mbox{\textrm{ otherwise}}
  \end{cases} \label{eq:final-density}
\end{equation}
Next, we determine the probability of a footprint as follows, 
\begin{multline}
P(F_{t}^{x,y \mid v_i}\textrm{ is Reached}) = \\ 
\int_{x-\frac{\triangle x}{2}}^{x+\frac{\triangle x}{2}} \int_{y-\frac{\triangle y}{2}}^{y+\frac{\triangle y}{2}} \int_{t-\frac{\triangle t}{2}}^{t+\frac{\triangle t}{2}} f(x,y,t \mid v_{i})dxdydt \\
\approx f(x,y,t \mid v_{i})\triangle x \triangle y \triangle t
\label{eq:p-reached}
\end{multline}
Where $\triangle x \triangle y$, and $\triangle t$ are the granularities with which we discretize the continuous spatiotemporal space of footprints. Equation \ref{eq:p-reached} may then be used in Eq. \ref{eq:p_lambda} and Eq. \ref{eq:p_zeta} to yield $P(\lambda)$ and $P(\zeta)$.

%%%%%%%%%%%%%%%%%%%%%%%%%%%%%%%%%%%%%%%%%%%%%%%%%%%%%%%%%%%%%%%%%%%%%%%%%%%%%%%%
\section{Evaluation Of Safety Metric} \label{s:eval}

We evaluate this approach by testing how well it identifies unsafe system-level outcomes when running an intentionally crippled prediction system and comparing this against an L2 error baseline. We used a cell size of $\triangle x = \triangle y = 0.5m$ and path-relative grid dimensions $30m\times10m$. We used $\triangle t = 0.3s$, a triangular distribution maxing out at $\pm15$ degrees for $f_{\theta}(\theta)$, and a Gaussian truncated at $\pm 3 m/s^2$ for $f_a(a)$. Additionally, we used the stricter $e'(F_{t}^{b})$ from Eq. \ref{eq:modified_eFt} in Eq. \ref{eq:p_lambda} (we did not observe significant difference in outcomes using $e'(F_{t}^{b})$ rather than $e(F_{t}^{b})$). We also set $t_1 = H-2$ in Eq. \ref{eq:protection}.

We took 99 challenging real-world scenarios encountered during manual data collection. Across all of these, there were several ``unpredictable'' pedestrians or cyclists. Human labelers tagged 145 of these as actors of interest (AOIs) because they were likely to interact with the ego-vehicle. Examples included people playing football in the middle of a busy intersection, panhandling, etc.

We then deployed an intentionally crippled motion prediction system (in which we curtailed trajectory lengths) to test this metric's efficacy at identifying unsafe scenarios. After integrating into a self-driving stack, we simulated all 99 scenarios (using a high-fidelity simulator in which simulated actors mimic their real-world counterparts' behavior). Eighteen scenarios produced unsafe outcomes, involving 27 AOIs (``unsafe AOIs''). We captured each actor's worst $P(\lambda_{\mathit{actor}})$ during its time in the spatial region of interest (from Fig. \ref{fig:sro_grid}). Next, we did the same with L2 error at 3s for actors in the spatial region of interest. We then separately ordered actors by descending $P(\lambda_{\mathit{actor}})$ and L2 error (larger values are worse for both).

\subsubsection{Global Signal-To-Noise Ratio}
We measure the signal-to-noise ratio by calculating the fraction of genuine violations in the top $N$ given an ordering of instances. First, we ordered all instances by both metrics separately. Then we filtered to the ``unsafe AOIs" flagged by the simulator and plotted histograms of their rankings (Fig. \ref{unsafe_aoi_dist}). When ordered by descending $P(\lambda_{\mathit{actor}})$, the ``unsafe AOIs" rank higher (farther left), suggesting superior SNR. Additionally, Fig. \ref{unsafe_aoi_snr}, which plots SNR against $N$ shows that the $P(\lambda_{\mathit{actor}})$ SNR is 40\% for $N=10$, 35\% for $N=20$ and so on, significantly above L2 error. 

Next, we examined orderings of \textit{all} AOI actors (not just ``unsafe" ones). As can be seen in Fig. \ref{all_aoi_dist}, the $P(\lambda_{\mathit{actor}})$ is better at delineating AOIs than L2 error (since more density is concentrated left). The $P(\lambda_{\mathit{actor}})$ SNR is also much greater; 60\% for $N=10$. 

\subsubsection{Actor Of Interest Sensitivity} 
We also considered how well the two metrics rank an AOI within a given scenario relative to other non-AOI actors encountered. As can be seen from Fig. \ref{unsafe_aoi_loglevel} and Fig. \ref{all_aoi_loglevel}, the $P(\lambda_{\mathit{actor}})$ metric is consistently capable of outperforming the L2 error baseline at ranking the AOI highly within a scenario.  

\subsubsection{Qualitative Instances}
\begin{figure}[H]
	\centering
	\begin{subfigure}{.5\linewidth}
		\centering
		 \captionsetup{justification=centering}
		\includegraphics[width=0.9in]{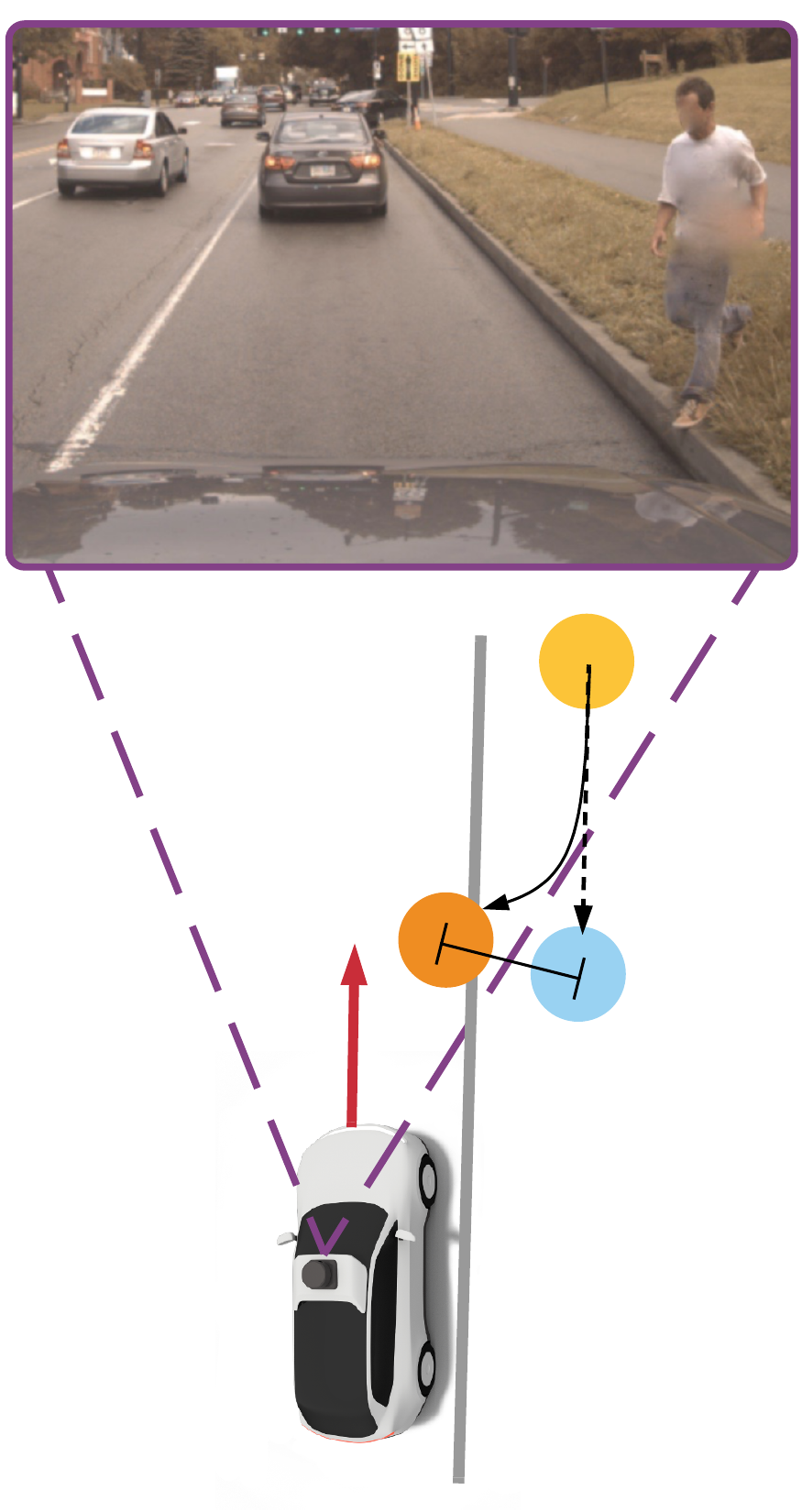}
		\caption{$P(\lambda_{actor})$ flags this as unsafe.}
		\label{fig:jaywalk1}
	\end{subfigure}%
	\begin{subfigure}{.5\linewidth}
		\centering
		 \captionsetup{justification=centering}
		\includegraphics[width=1in]{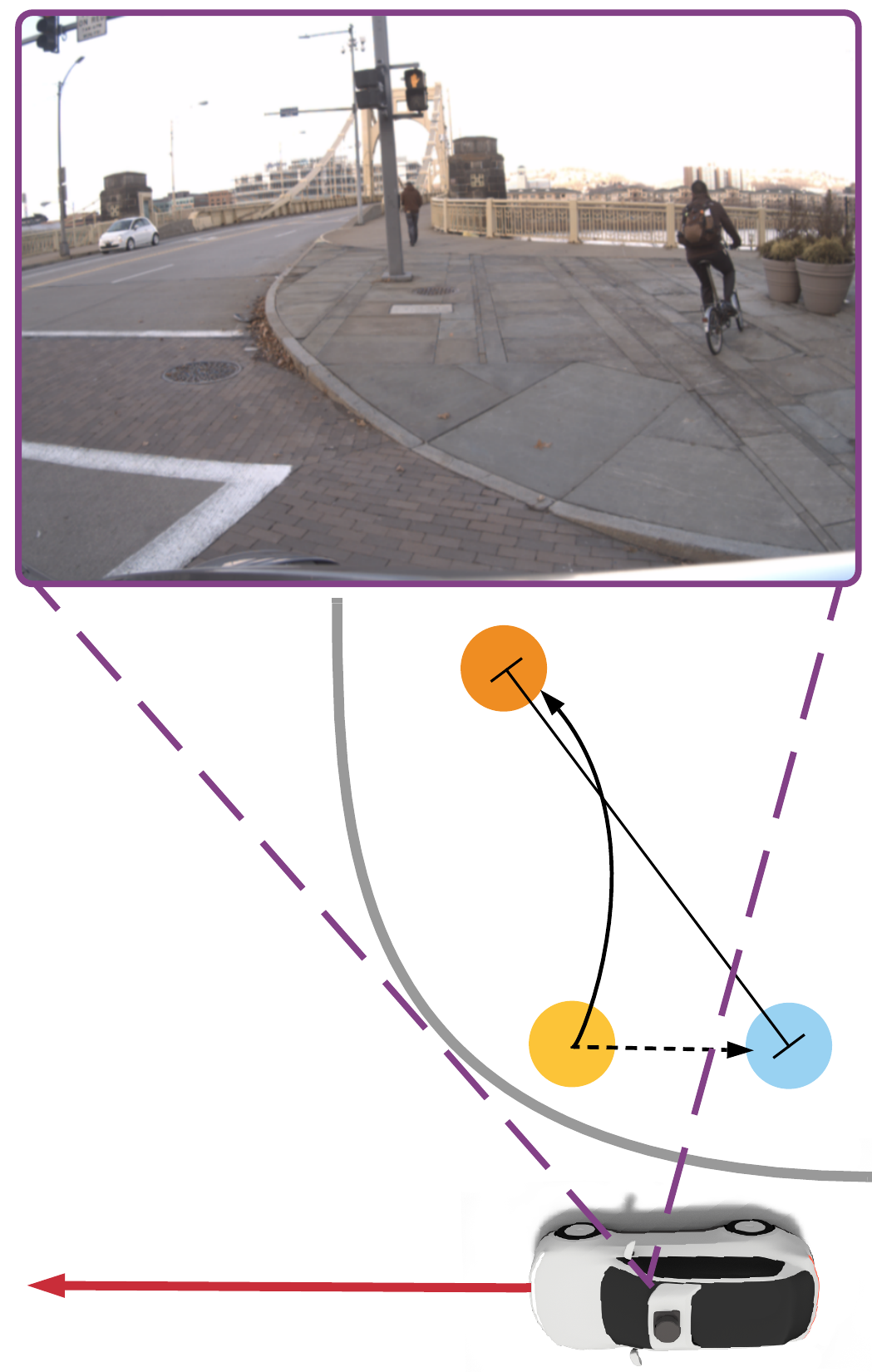}
		\caption{$P(\lambda_{actor})$ judges this as not unsafe.}
		\label{fig:pgh_bik}
	\end{subfigure}
	\caption{Ground-truth at $t+0$ s (yellow), $t + 3$ s (orange) and the predictions at $t + 3$ s (blue) for two real-world instances.}
\end{figure}

Consider Fig. \ref{fig:jaywalk1} which depicts a person stepping into the street. The artificially crippled predictions (blue) fail to predict this actor's ground-truth future motion (orange). Among pedestrians and cyclists, $P(\lambda_{\mathit{actor}})$ ranks this AOI 9th; when ordered by L2 error at 3s, it ranks 120th. The L2 error between proceeding down the sidewalk and stepping into the street is too slight. However, the safety consequence is large. 

Figure \ref{fig:pgh_bik} shows the opposite case. The prediction fails to anticipate the cyclist's off-road trajectory, and a large L2 error results. Ordered by L2, this instance ranks 3rd amongst cyclists and pedestrians while $P(\lambda_{\mathit{actor}})$  ranks it 108th, which is more appropriate since there is little chance of a risky interaction with this actor.
 
%%%%%%%%%%%%%%%%%%%%%%%%%%%%%%%%%%%%%%%%%%%%%%%%%%%%%%%%%%%%%%%%%%%%%%%%%%%%%%%%
\section{Conclusions and Future Work}

In this work, we demonstrated a method to directly translate the concepts of recall and precision into the self-driving specific metrics $P(\lambda)$ and $P(\zeta)$ reflecting safety and ride comfort. We proposed a general formulation that applies to \textit{any} ego-vehicle trajectory distribution or spatiotemporal grid representation. 

Next, we applied our formulation to a particular distribution of simple ego-vehicle trajectories  (beelines), which we make agnostic to road-geometry using the path-relative transformation. We are thus able to use the same set of beelines for \textit{any} potential scene, as long as a nominal path is available and the transformation is well defined.

Using a simulator, we demonstrated that our method shows a much better signal-to-noise ratio than displacement error at identifying risky events. This work evaluated the safety metric $P(\lambda)$; in future work, we plan to repeat the same for comfort $P(\zeta)$. 

%%%%%%%%%%%%%%%%%%%%%%%%%%%%%%%%%%%%%%%%%%%%%%%%%%%%%%%%%%%%%%%%%%%%%%%%%%%%%%%%

%First table of results
\begin{figure*}[h]
\begin{subfigure}{.475\linewidth}
\centering
\begin{tikzpicture}
  \begin{axis}
  [
    axis lines=center,
    ymax=8,
    ymin=0,
    xmin=-20,
    xmax=720,
    samples at={0,1,...,50},
    yticklabel style=
    {
      /pgf/number format/fixed,
      /pgf/number format/precision=1
    },
    ybar=0pt, 
    bar width=2,
    xlabel=$Rank$,
    ylabel=$\textrm{Instance Count}$, minor x tick num=0, 
    ]
   \addplot table [x=bin, y=bl, col sep=comma] {datax.dat};
   \addlegendentry{$P(\lambda_{\mathit{actor}})$}
   \addplot table [x=bin, y=l2, col sep=comma] {datax.dat};
   \addlegendentry{\textrm{L2 in ROI}}
  \end{axis}
\end{tikzpicture}
\caption{Histogram of rankings for unsafe AOIs.\newline}
\label{unsafe_aoi_dist}
\end{subfigure}
\begin{subfigure}{.475\linewidth}
\centering
\begin{tikzpicture}
  \begin{axis}
  [
   axis lines=center,
    ymax=1,
    ymin=0,
    xmin=-1,
    xmax=720,
    samples at={0,1,...,50},
    yticklabel style=
    {
      /pgf/number format/fixed,
      /pgf/number format/precision=1
    },
    xlabel=$Rank$,
    ylabel=$\textrm{Signal-To-Noise Ratio}$, minor x tick num=0, 
    ]
   \addplot table [x=bin,y=bl_snr, col sep=comma] {datax.dat};
   \addlegendentry{$P(\lambda_{\mathit{actor}})$}
   \addplot table [x=bin,y=l2_snr, col sep=comma] {datax.dat};
   \addlegendentry{\textrm{L2 in ROI}}
  \end{axis}
\end{tikzpicture}
\caption{Percentage of actors in each bin which are unsafe AOIs.\newline}
\label{unsafe_aoi_snr}
\end{subfigure} %\vspace{5mm}

\begin{subfigure}{\linewidth}
\begin{tabularx}{\linewidth}{*{6}{>{\centering\arraybackslash}X}}
  \hline
                           & $\mathbf{P(\lambda_{\mathit{actor}})}$ & \textbf{L2 in ROI} &                                & $\mathbf{P(\lambda_{\mathit{actor}})}$ & \textbf{L2 in ROI} \\
   \hline
    \textbf{Top 1} &  56\% (15/27)          &  7\% (2/27)           &  \textbf{Top  5\%}   &  59\% (16/27)         &  7\% (2/27)         \\     
    \textbf{Top 2} &  78\% (21/27).         &  30\% (8/27)         &  \textbf{Top  10\%} &  63\% (17/27)         &  15\% (4/27)       \\   
    \textbf{Top 3} &  89\% (24/27)          &  37\% (10/27)       &  \textbf{Top  25\%} &  85\% (23/27)         &   22\% (6/27)      \\   
    \textbf{Top 4} &  96\% (26/27)          &  56\% (15/27)       &  \textbf{Top  33\%} &  89\% (24/27)         &   26\% (7/27)      \\   
    \textbf{Top 5} &  96\% (26/27)          &  67\% (18/27)       &  \textbf{Top  50\%} &  96\% (26/27)         &    63\% (17/27)       \\   
   \hline
\end{tabularx}
\caption{Percentage of scenes where unsafe AOI ranked in Top K within scene (left) or Top Q\textsuperscript{th} percentile within scene (right).\newline}
\label{unsafe_aoi_loglevel}
\end{subfigure} %\vspace{5mm}
\caption{Statistics for known ``unsafe'' AOIs}
\end{figure*}

%Second table of results
\begin{figure*}
\begin{subfigure}{.475\linewidth}
\centering
\begin{tikzpicture}
  \begin{axis}
  [
    axis lines=center,
    ymax=8,
    ymin=0,
    xmin=-20,
    xmax=720,
    samples at={0,1,...,50},
    yticklabel style=
    {
      /pgf/number format/fixed,
      /pgf/number format/precision=1
    },
    ybar=0pt, 
    bar width=2,
    xlabel=$Rank$,
    ylabel=$\textrm{Instance Count}$, minor x tick num=0, 
    ]
   \addplot table [x=bin, y=bl, col sep=comma] {datax_all.dat};
   \addlegendentry{$P(\lambda_{\mathit{actor}})$}
   \addplot table [x=bin, y=l2, col sep=comma] {datax_all.dat};
   \addlegendentry{\textrm{L2 in ROI}}
  \end{axis}
\end{tikzpicture}
\caption{Histogram of rankings for all AOIs. \newline}
\label{all_aoi_dist}
\end{subfigure}
\begin{subfigure}{.475\linewidth}
\centering
\begin{tikzpicture}
  \begin{axis}
  [
    axis lines=center,
    ymax=1,
    ymin=0,
    xmin=-1,
    xmax=720,
    samples at={0,1,...,50},
    yticklabel style=
    {
      /pgf/number format/fixed,
      /pgf/number format/precision=1
    },
    xlabel=$Rank$,
    ylabel=$\textrm{Signal-To-Noise Ratio}$, minor x tick num=0, 
    ]
   \addplot table [x=bin,y=bl_snr, col sep=comma] {datax_all.dat};
   \addlegendentry{$P(\lambda_{\mathit{actor}})$}
   \addplot table [x=bin,y=l2_snr, col sep=comma] {datax_all.dat};
   \addlegendentry{\textrm{L2 in ROI}}
  \end{axis}
\end{tikzpicture}
\caption{Percentage of actors in each bin which are AOIs. \newline}
\label{all_aoi_snr}
\end{subfigure} %\vspace{5mm}
\begin{subfigure}{\linewidth}
\begin{tabularx}{\linewidth}{*{6}{>{\centering\arraybackslash}X}}
  \hline
                           & $\mathbf{P(\lambda_{\mathit{actor}})}$ & \textbf{L2 in ROI} &                                & $\mathbf{P(\lambda_{\mathit{actor}})}$ & \textbf{L2 in ROI} \\
   \hline
    \textbf{Top 1} & 41\% (60/145)             &  9\% (13/145)         &  \textbf{Top  5\%}   &  43\% (62/145)            &  9\% (13/145)         \\     
    \textbf{Top 2} &  69\% (100/145)          &  25\% (36/145)       &  \textbf{Top  10\%} &  52\% (76/145)            &  13\% (19/145)       \\   
    \textbf{Top 3} &  81\% (118/145)          &  35\% (51/145)       &  \textbf{Top  25\%} &  75\% (109/145)          &   21\% (30/145)      \\   
    \textbf{Top 4} &  89\% (129/145)          &  51\% (74/145)       &  \textbf{Top  33\%} &  79\% (114/145)          &   25\% (36/145)       \\   
    \textbf{Top 5} &  91\% (132/145)          &  59\% (85/145)       &  \textbf{Top  50\%} &  91\% (132/145)          &   54\% (79/145)       \\   
   \hline
\end{tabularx}
\caption{Percentage of scenes where AOI ranked in Top K actors within scene (left) or Top Q\textsuperscript{th} percentile within scene (right). \newline}
\label{all_aoi_loglevel}
\end{subfigure} %\vspace{5mm}
\caption{Statistics for all AOIs}
\end{figure*}

\bibliographystyle{IEEEtran}
\bibliography{biblio}

\begin{thebibliography}{10}
\providecommand{\url}[1]{#1}
\csname url@samestyle\endcsname
\providecommand{\newblock}{\relax}
\providecommand{\bibinfo}[2]{#2}
\providecommand{\BIBentrySTDinterwordspacing}{\spaceskip=0pt\relax}
\providecommand{\BIBentryALTinterwordstretchfactor}{4}
\providecommand{\BIBentryALTinterwordspacing}{\spaceskip=\fontdimen2\font plus
\BIBentryALTinterwordstretchfactor\fontdimen3\font minus
  \fontdimen4\font\relax}
\providecommand{\BIBforeignlanguage}[2]{{%
\expandafter\ifx\csname l@#1\endcsname\relax
\typeout{** WARNING: IEEEtran.bst: No hyphenation pattern has been}%
\typeout{** loaded for the language `#1'. Using the pattern for}%
\typeout{** the default language instead.}%
\else
\language=\csname l@#1\endcsname
\fi
#2}}
\providecommand{\BIBdecl}{\relax}
\BIBdecl

\bibitem{pellegrini2009}
S.~Pellegrini, A.~Ess, K.~Schindler, and L.~Van~Gool, ``You'll never walk
  alone: Modeling social behavior for multi-target tracking,'' in \emph{2009
  IEEE 12th International Conference on Computer Vision}.\hskip 1em plus 0.5em
  minus 0.4em\relax IEEE, 2009, pp. 261--268.

\bibitem{trajnet}
A.~Sadeghian, V.~Kosaraju, A.~Gupta, S.~Savarese, and A.~Alahi, ``Trajnet:
  Towards a benchmark for human trajectory prediction,'' \emph{arXiv preprint},
  2018.

\bibitem{red-predictor}
S.~Becker, R.~Hug, W.~Hubner, and M.~Arens, ``Red: A simple but effective
  baseline predictor for the trajnet benchmark,'' in \emph{Proceedings of the
  European Conference on Computer Vision (ECCV)}, 2018, pp. 0--0.

\bibitem{djuric2018}
N.~Djuric, V.~Radosavljevic, H.~Cui, T.~Nguyen, F.-C. Chou, T.-H. Lin, and
  J.~Schneider, ``Short-term motion prediction of traffic actors for autonomous
  driving using deep convolutional networks,'' \emph{arXiv preprint
  arXiv:1808.05819}, 2018.

\bibitem{becker2019rnn}
S.~Becker, R.~Hug, W.~H{\"u}bner, and M.~Arens, ``An rnn-based imm filter
  surrogate,'' in \emph{Scandinavian Conference on Image Analysis}.\hskip 1em
  plus 0.5em minus 0.4em\relax Springer, 2019, pp. 387--398.

\bibitem{nuscenes-challenge}
\BIBentryALTinterwordspacing
H.~Caesar, V.~Bankiti, A.~H. Lang, S.~Vora, V.~E. Liong, Q.~Xu, A.~Krishnan,
  Y.~Pan, G.~Baldan, and O.~Beijbom. (2020) nuscenes prediction task. [Online].
  Available: \url{https://www.nuscenes.org/prediction}
\BIBentrySTDinterwordspacing

\bibitem{argoverse-challenge}
\BIBentryALTinterwordspacing
M.-F. Chang, J.~Lambert, P.~Sangkloy, J.~Singh, S.~Bak, A.~Hartnett, D.~Wang,
  P.~Carr, S.~Lucey, D.~Ramanan \emph{et~al.} (2019) Argoverse motion
  forecasting competition. [Online]. Available:
  \url{https://evalai.cloudcv.org/web/challenges/challenge-page/454/evaluation}
\BIBentrySTDinterwordspacing

\bibitem{desire}
N.~Lee, W.~Choi, P.~Vernaza, C.~B. Choy, P.~H. Torr, and M.~Chandraker,
  ``Desire: Distant future prediction in dynamic scenes with interacting
  agents,'' in \emph{Proceedings of the IEEE Conference on Computer Vision and
  Pattern Recognition}, 2017, pp. 336--345.

\bibitem{cui2019}
H.~Cui, V.~Radosavljevic, F.-C. Chou, T.-H. Lin, T.~Nguyen, T.-K. Huang,
  J.~Schneider, and N.~Djuric, ``Multimodal trajectory predictions for
  autonomous driving using deep convolutional networks,'' in \emph{2019
  International Conference on Robotics and Automation (ICRA)}.\hskip 1em plus
  0.5em minus 0.4em\relax IEEE, 2019, pp. 2090--2096.

\bibitem{rupprecht2017}
C.~Rupprecht, I.~Laina, R.~DiPietro, M.~Baust, F.~Tombari, N.~Navab, and G.~D.
  Hager, ``Learning in an uncertain world: Representing ambiguity through
  multiple hypotheses,'' in \emph{Proceedings of the IEEE International
  Conference on Computer Vision}, 2017, pp. 3591--3600.

\bibitem{safecritic}
T.~van~der Heiden, N.~S. Nagaraja, C.~Weiss, and E.~Gavves, ``Safecritic:
  Collision-aware trajectory prediction,'' \emph{arXiv preprint
  arXiv:1910.06673}, 2019.

\bibitem{lyft-challenge}
\BIBentryALTinterwordspacing
J.~Houston, G.~Zuidhof, L.~Bergamini, Y.~Ye, A.~Jain, S.~Omari, V.~Iglovikov,
  and P.~Ondruska. (2020) Lyft motion prediction for autonomous vehicles.
  [Online]. Available:
  \url{https://github.com/lyft/l5kit/blob/master/competition.md}
\BIBentrySTDinterwordspacing

\bibitem{precog}
N.~Rhinehart, R.~McAllister, K.~Kitani, and S.~Levine, ``Precog: Prediction
  conditioned on goals in visual multi-agent settings,'' in \emph{Proceedings
  of the IEEE International Conference on Computer Vision}, 2019, pp.
  2821--2830.

\bibitem{schulz2018interaction}
J.~Schulz, C.~Hubmann, J.~L{\"o}chner, and D.~Burschka, ``Interaction-aware
  probabilistic behavior prediction in urban environments,'' in \emph{2018
  IEEE/RSJ International Conference on Intelligent Robots and Systems
  (IROS)}.\hskip 1em plus 0.5em minus 0.4em\relax IEEE, 2018, pp. 3999--4006.

\bibitem{multipath}
Y.~Chai, B.~Sapp, M.~Bansal, and D.~Anguelov, ``Multipath: Multiple
  probabilistic anchor trajectory hypotheses for behavior prediction,''
  \emph{arXiv preprint arXiv:1910.05449}, 2019.

\bibitem{jain2019}
A.~Jain, S.~Casas, R.~Liao, Y.~Xiong, S.~Feng, S.~Segal, and R.~Urtasun,
  ``Discrete residual flow for probabilistic pedestrian behavior prediction,''
  \emph{arXiv preprint arXiv:1910.08041}, 2019.

\bibitem{chauffeurnet}
M.~Bansal, A.~Krizhevsky, and A.~Ogale, ``Chauffeurnet: Learning to drive by
  imitating the best and synthesizing the worst,'' \emph{arXiv preprint
  arXiv:1812.03079}, 2018.

\bibitem{faf}
W.~Luo, B.~Yang, and R.~Urtasun, ``Fast and furious: Real time end-to-end 3d
  detection, tracking and motion forecasting with a single convolutional net,''
  in \emph{Proceedings of the IEEE conference on Computer Vision and Pattern
  Recognition}, 2018, pp. 3569--3577.

\end{thebibliography}
\end{document}